\documentclass[conference]{IEEEtran}
\IEEEoverridecommandlockouts
\usepackage{cite}
\usepackage{amsmath,amssymb,amsfonts}
\usepackage{algorithmic}
\usepackage{graphicx}
\usepackage{textcomp}
\usepackage{hyperref}
\usepackage{academicons}
\usepackage{xcolor}

\usepackage{epsfig}

\usepackage{epsfig}
\usepackage{subcaption}
\usepackage{calc}
\usepackage{amssymb}
\usepackage{amstext}
\usepackage{amsthm}
\usepackage{multicol}
\usepackage{pslatex}
\usepackage{apalike}
\usepackage[ruled,vlined, linesnumbered]{algorithm2e}
\usepackage[bottom]{footmisc}
\usepackage{multirow}
\usepackage{nicefrac}

\usepackage{pgf}
\usepackage{ulem}
\usepackage{hyperref}
\usepackage{cleveref}
\usepackage{amsfonts}
\usepackage{textcomp}
\usepackage{xcolor}

\usepackage{float}
\usepackage{longtable}
\usepackage{wrapfig}
\usepackage{tikz}
\usepackage{standalone}
\usepackage{bm}
\usepackage{bbm}

\definecolor{LightCyan}{rgb}{0.88,1,1}
\definecolor{chalkblue}{rgb}{0.671, 0.871, 0.902}
\definecolor{chalkpurple}{rgb}{0.796, 0.667,0.796}
\definecolor{chalkyellow}{rgb}{1.0, 1.0, 0.71}
\definecolor{chalkorange}{rgb}{1.0, 0.8, 0.714}
\definecolor{chalkpink}{rgb}{0.953, 0.69,0.765}

\definecolor{bohored}{RGB}{209, 133, 119}
\definecolor{bohogreen}{RGB}{130, 157, 136}
\definecolor{bohoblue}{RGB}{160, 174, 189}
\definecolor{bohoyellow}{RGB}{225, 180, 105}

\definecolor{mellowpurple}{RGB}{213, 104, 171}
\definecolor{mellowpink}{RGB}{250, 200, 220}
\definecolor{mellowgreen}{RGB}{90, 146, 91 }
\definecolor{mellowblue}{RGB}{177, 222, 220}
\definecolor{mellowyellow}{RGB}{ 254, 192, 93}
\definecolor{melloworange}{RGB}{226, 117, 76}

\definecolor{sunsetpink}{RGB}{162, 59, 85}
\definecolor{applegreen}{rgb}{0.55, 0.71, 0.0}
\definecolor{airforceblue}{rgb}{0.36, 0.54, 0.66}
\definecolor{amethyst}{rgb}{0.6, 0.4, 0.8}
\definecolor{antiquefuchsia}{rgb}{0.57, 0.36, 0.51}
\definecolor{aquamarine}{rgb}{0.5, 1.0, 0.83}
\definecolor{asparagus}{rgb}{0.53, 0.66, 0.42}
\definecolor{babyblue}{rgb}{0.54, 0.81, 0.94}
\definecolor{babyblueeyes}{rgb}{0.63, 0.79, 0.95}
\definecolor{babypink}{rgb}{0.96, 0.76, 0.76}
\definecolor{darkseagreen}{rgb}{0.56, 0.74, 0.56}
\definecolor{flavescent}{rgb}{0.97, 0.91, 0.56}
\definecolor{grannysmithapple}{rgb}{0.66, 0.89, 0.63}
\definecolor{pastelorange}{rgb}{1.0, 0.7, 0.28}
\definecolor{pastelmagenta}{rgb}{0.96, 0.6, 0.76}
\definecolor{richelectricblue}{rgb}{0.03, 0.57, 0.82}
\definecolor{rosevale}{rgb}{0.67, 0.31, 0.32}
\definecolor{sandstorm}{rgb}{0.93, 0.84, 0.25}

\definecolor{veryperi}{rgb}{10,99,255}
\definecolor{orchidbloom}{rgb}{194,166,245}

\definecolor{popcorn}{cmyk}{3,13,53,0}
\definecolor{bubblegum}{cmyk}{3,67,25,0}
\definecolor{orchidbloom}{cmyk}{24,35,4,0}
\definecolor{daffodil}{cmyk}{0,29,76,0}
\definecolor{poinciana}{cmyk}{14,89,92,4}
\definecolor{harborblue}{cmyk}{87,37,44,27}
\definecolor{Cascade}{cmyk}{57,4,36,0}
\definecolor{spunsugar}{cmyk}{33,1,7,0}
\definecolor{coccamocha}{cmyk}{36,48,55,34}
\definecolor{fragilesprout}{cmyk}{34,13,93,1}
\definecolor{supersonic}{cmyk}{95,64,10,1}

\newcommand{\R}{\mathbb{R}}
\newcommand{\N}{\mathbb{N}}

\newcommand{\argmax}{\operatorname{argmax}}

\renewcommand{\l}{\ell}

\newcommand{\ab}{^{\rm A, B}}

\newcommand{\pd}{{p_d}}
\newcommand{\U}{{\mathcal{U}}}

\newcommand{\rk}{r^{(k)}}

\newcommand{\uk}{u^{(k)}}

\newcommand{\ukk}{u^{(k+1)}}

\newcommand{\lmin}{\lambda_{\min}}
\newcommand{\lmax}{\lambda_{\max}}
\newcommand{\bn}{\operatorname{bn}}

\newcommand{\Bk}{B^{(k)}}

\newcommand{\relu}{\operatorname{ReLU}}

\newcommand\orcid[2]{
\hspace*{-1.5mm}$^\text{, #2}$\thanks{$^\text{#2}$https://orcid.org/#1}\hspace*{-1mm}
}

\def\BibTeX{{\rm B\kern-.05em{\sc i\kern-.025em b}\kern-.08em
    T\kern-.1667em\lower.7ex\hbox{E}\kern-.125emX}}
\begin{document}

\title{Poly-MgNet: Polynomial Building Blocks in Multigrid-Inspired ResNets
\thanks{This work is partially supported by the German Federal Ministry for Economic Affairs and Climate Action, within the project “KI Delta Learning” (grant no.\ 19A19013Q). M.R.\ acknowledges support by the German Federal Ministry of Education and Research within the junior research group project “UnrEAL” (grant no.\ 01IS22069). The contribution of K.K.\ is partially funded by the European Union’s HORIZON MSCA Doctoral Networks programme project AQTIVATE (grant no.\ 101072344).}
}

\author{
\IEEEauthorblockN{Antonia van Betteray$^{\text{1}}$\orcid{0000-0002-2338-1753}{a},
Matthias Rottmann$^{\text{1}}$\orcid{0000-0003-3840-0184}{b} and
Karsten Kahl$^{\text{1}}$\orcid{0000-0003-3840-0184}{c}}
\IEEEauthorblockA{
\textit{$^{1}$IZMD, University of Wuppertal, Germany} \\
\{vanbetteray, rottmann, kkahl\}@uni-wuppertal.de\\} }

\maketitle

\begin{abstract}
The structural analogies of ResNets and Multigrid (MG) methods such as common building blocks like convolutions and poolings where already pointed out by He et al.\ in 2016. Multigrid methods are used in the context of scientific computing for solving large sparse linear systems arising from partial differential equations. MG methods particularly rely on two main concepts: smoothing and residual restriction / coarsening. Exploiting these analogies, He and Xu developed the MgNet framework, which integrates MG schemes into the design of ResNets. In this work, we introduce a novel neural network building block inspired by polynomial smoothers from MG theory. Our polynomial block from an MG perspective naturally extends the MgNet framework to Poly-Mgnet and at the same time reduces the number of weights in MgNet. We present a comprehensive study of our polynomial block, analyzing the choice of initial coefficients, the polynomial degree, the placement of activation functions, as well as of batch normalizations. 
{Our results demonstrate that constructing (quadratic) polynomial building blocks based on real and imaginary polynomial roots enhances Poly-MgNet's capacity in terms of accuracy.}
Furthermore, our approach achieves an improved trade-off of model accuracy and number of weights compared to ResNet as well as compared to specific configurations of MgNet.
\end{abstract}


\section{{Introduction}}
\label{sec:introduction}
Deep convolutional neural networks (CNNs) are state-of-the-art methods for image classification tasks~\cite{krizhevsky_imagenet_2012,russakovsky_imagenet_2015,he_delving_2015,Dosovitskiy2020}. Especially ResNets~\cite{he_deep_2016,he_identity_2016,liu2022convnet} have become increasingly popular, as they successfully overcome the vanishing gradient problem~\cite{Glorot2010UnderstandingTD}.

\begin{figure}[t]
\centering
\scalebox{.75}{
    \hfill
    \begin{subfigure}[t]{0.26\linewidth}
    \centering



\tikzstyle{block} = [draw, fill=white, rectangle, minimum height=1.5em, minimum width=7.5em,inner sep=0pt]
\begin{tikzpicture}

\coordinate (inp) at (0, 0.8);
\node [block, fill=bohoyellow] (A1) at (0,0) {$A_{\l}^i$};
\node [block, fill=bohogreen] (B1) at (0,-1.5) {$B_{\l}^{i}$};

\coordinate [] (p1-1) at (0, 0.5) {};
\coordinate [] (p2-1) at (1.5, 0.5) {};
\coordinate [] (p3-1) at (1.5, -2.25) {};

\coordinate [] (f1-1) at (0, 0.65) {};
\coordinate [] (f2-1) at (1.65, 0.65) {};
\coordinate [] (f3-1) at (1.65, -.75) {};

\node [name=conc, draw, circle, minimum size=4pt, inner sep = 0, node distance=1cm] at (0, -2.25) {$+$};

\draw[- latex] (inp) -- (A1);
\draw[- latex] (A1) -- (B1) node[midway, left] {$\sigma$};
\draw[- latex] (B1) -- (conc) node[midway, left] {$\sigma$};

\draw[- latex] (p1-1) -- (p2-1) -- (p3-1) -- (conc);

\node [block, fill= bohored] (A2) at (0,-3) {$A_{\l}^{i+1}$};
\node [block, fill = bohoblue] (B2) at (0, -4.5) {$B_{\l}^{i+1}$};

\coordinate [] (p1-2) at (0, -2.5) {};
\coordinate [] (p2-2) at (1.5, -2.5) {};
\coordinate [] (p3-2) at (1.5, -5.25) {};

\coordinate [] (f3-2) at (1.65, -3.75) {};
\coordinate [] (f-out) at (1.65, -5.8) {};

\node [name=conc2, draw, circle, minimum size=4pt, inner sep = 0, node distance=1cm] at (0, -5.25) {$+$};
\coordinate (out2) at (0, -5.8);

 \draw[-latex] (p1-2) -- (p2-2) -- (p3-2) -- (conc2);

\draw[- latex] (conc) -- (A2);
\draw[- latex] (A2) -- (B2) node[midway, left] {$\sigma$};
\draw[- latex] (B2) -- (conc2) node[midway, left] {$\sigma$};
\draw[- latex] (conc2) -- (out2);

\end{tikzpicture}
        \caption{residual}\label{subfig:resnet}
    \end{subfigure}
    \hfill
    \begin{subfigure}[t]{0.26\linewidth}
    \centering
        \tikzstyle{block} = [draw, fill=white, rectangle, minimum height=1.5em, minimum width=7.5em,inner sep=0pt]
\begin{tikzpicture}

\coordinate (inp) at (0, 0.8);
\node [block, fill=bohoyellow] (A1) at (0,0) {$A_{\l}$};
\node [block, fill=bohogreen] (B1) at (0,-1.5) {$B_{\l}^{(k)}$};

\coordinate [] (p1-1) at (0, 0.5) {};
\coordinate [] (p2-1) at (1.5, 0.5) {};
\coordinate [] (p3-1) at (1.5, -2.25) {};

\coordinate [] (f1-1) at (0, 0.65) {};
\coordinate [] (f2-1) at (1.65, 0.65) {};
\coordinate [] (f3-1) at (1.65, -.75) {};

\node [name=res1, draw, circle, minimum size=1pt, inner sep = 0] at(0, -.75) {$-$};
\node [name=conc, draw, circle, minimum size=4pt, inner sep = 0, node distance=1cm] at (0, -2.25) {$+$};

\draw[-latex] (inp) -- (A1);
\draw[-latex] (A1) -- (res1);
\draw[-latex] (res1) -- (B1) node[pos=.35, left] {$\sigma$};
\draw[-latex] (B1) -- (conc) node[midway, left] {$\sigma$};

\draw[-latex] (p1-1) -- (p2-1) -- (p3-1) -- (conc);
\draw[-latex, dashed] (f1-1) -- (f2-1) -- (f3-1) -- (res1);

\node [block, fill= bohoyellow] (A2) at (0,-3) {$A_{\l}$};
\node [block, fill = bohored] (B2) at (0, -4.5) {$B_{\l}^{(k+1)}$};

\coordinate [] (p1-2) at (0, -2.5) {};
\coordinate [] (p2-2) at (1.5, -2.5) {};
\coordinate [] (p3-2) at (1.5, -5.25) {};

\coordinate [] (f1-2) at (0, -2) {};
\coordinate [] (f2-2) at (1.65, -2) {};
\coordinate [] (f3-2) at (1.65, -3.75) {};
\coordinate [] (f-out) at (1.65, -5.8) {};

\node [name=res2, draw, circle, minimum size=1pt, inner sep = 0] at(0, -3.75) {$-$};
\node [name=conc2, draw, circle, minimum size=4pt, inner sep = 0, node distance=1cm] at (0, -5.25) {$+$};
\coordinate (out) at (0, -5.8);

\draw[-latex] (p1-2) -- (p2-2) -- (p3-2) -- (conc2);
\draw[-latex, dashed] (f3-1) -- (f2-2) -- (f3-2) -- (res2);
\draw[-latex, dashed] (f3-2) -- (f-out); 

\draw[-latex] (conc) -- (A2);
\draw[-latex] (A2) -- (res2);
\draw[-latex] (res2) -- (B2) node[pos=.35, left] {$\sigma$};
\draw[-latex] (B2) -- (conc2) node[midway, left] {$\sigma$};
\draw[-latex] (conc2) -- (out);

\end{tikzpicture}
        \caption{non-stationary}\label{subfig:mgneti}
    \end{subfigure}
    \hfill
    \begin{subfigure}[t]{0.26\linewidth}
    \centering

\tikzstyle{block} = [draw, fill=white, rectangle, minimum height=1.5em, minimum width=7.5em,inner sep=0pt]
\begin{tikzpicture}

\coordinate (inp) at (0, 0.8);
\node [block, fill=bohoyellow] (A1) at (0,0) {$A_{\l}$};
\node [block, fill=bohogreen] (B1) at (0,-1.5) {$B_{\l}$};

\coordinate [] (p1-1) at (0, 0.5) {};
\coordinate [] (p2-1) at (1.5, 0.5) {};
\coordinate [] (p3-1) at (1.5, -2.25) {};

\coordinate [] (f1-1) at (0, 0.65) {};
\coordinate [] (f2-1) at (1.65, 0.65) {};
\coordinate [] (f3-1) at (1.65, -.75) {};

\node [name=res1, draw, circle, minimum size=1pt, inner sep = 0] at(0, -.75) {$-$};
\node [name=conc, draw, circle, minimum size=4pt, inner sep = 0, node distance=1cm] at (0, -2.25) {$+$};

\draw[-latex] (inp) -- (A1);
\draw[-latex] (A1) -- (res1);
\draw[-latex] (res1) -- (B1) node[pos=.35, left] {$\sigma$};
\draw[-latex] (B1) -- (conc) node[midway, left] {$\sigma$};

\draw[-latex] (p1-1) -- (p2-1) -- (p3-1) -- (conc);
\draw[-latex, dashed] (f1-1) -- (f2-1) -- (f3-1) -- (res1);

\node [block, fill= bohoyellow] (A2) at (0,-3) {$A_{\l}$};
\node [block, fill = bohogreen] (B2) at (0, -4.5) {$B_{\l}$};

\coordinate [] (p1-2) at (0, -2.5) {};
\coordinate [] (p2-2) at (1.5, -2.5) {};
\coordinate [] (p3-2) at (1.5, -5.25) {};

\coordinate [] (f1-2) at (0, -2) {};
\coordinate [] (f2-2) at (1.65, -2) {};
\coordinate [] (f3-2) at (1.65, -3.75) {};
\coordinate [] (f-out) at (1.65, -5.8) {};

\node [name=res2, draw, circle, minimum size=1pt, inner sep = 0] at(0, -3.75) {$-$};
\node [name=conc2, draw, circle, minimum size=4pt, inner sep = 0, node distance=1cm] at (0, -5.25) {$+$};
\coordinate (out) at (0, -5.8);

\draw[-latex] (p1-2) -- (p2-2) -- (p3-2) -- (conc2);
\draw[-latex, dashed] (f3-1) -- (f2-2) -- (f3-2) -- (res2);
\draw[-latex, dashed] (f3-2) -- (f-out); 

\draw[-latex] (conc) -- (A2);
\draw[-latex] (A2) -- (res2);
\draw[-latex] (res2) -- (B2) node[pos=.35, left] {$\sigma$};
\draw[-latex] (B2) -- (conc2) node[midway, left] {$\sigma$};
\draw[-latex] (conc2) -- (out);

\end{tikzpicture}
        \caption{stationary}\label{subfig:mgnet}
    \end{subfigure}
        \hfill
    \begin{subfigure}[t]{0.26\linewidth}
    \centering
        \tikzstyle{block} = [draw, fill=white, rectangle, minimum height=1.5em, minimum width=7.5em,inner sep=0pt]

\tikzstyle{poly} = [draw, fill=white, rectangle, minimum height=1.5em, minimum width=3.5em,inner sep=0pt]
\begin{tikzpicture}

\coordinate (inp) at (0, 0.8);
\node [block, fill=bohoyellow] (A1) at (0,0) {$A_{\l}$};
\node [poly, fill=bohogreen] (B1) at (0,-1.5) {$\alpha_{\l}^{(k+1)}$};

\coordinate [] (p1-1) at (0, 0.5) {};
\coordinate [] (p2-1) at (1.5, 0.5) {};
\coordinate [] (p3-1) at (1.5, -2.25) {};

\coordinate [] (f1-1) at (0, 0.65) {};
\coordinate [] (f2-1) at (1.65, 0.65) {};
\coordinate [] (f3-1) at (1.65, -.75) {};

\node [name=res1, draw, circle, minimum size=1pt, inner sep = 0] at(0, -.75) {$-$};
\node [name=conc, draw, circle, minimum size=4pt, inner sep = 0, node distance=1cm] at (0, -2.25) {$+$};

\draw[-latex] (inp) -- (A1);
\draw[-latex] (A1) -- (res1);
\draw[-latex] (res1) -- (B1) node[pos=.35, left] {$\sigma$};
\draw[-latex] (B1) -- (conc) node[midway, left] {$\sigma$};

\draw[-latex] (p1-1) -- (p2-1) -- (p3-1) -- (conc);
\draw[-latex, dashed] (f1-1) -- (f2-1) -- (f3-1) -- (res1);

\node [block, fill= bohoyellow] (A2) at (0,-3) {$A_{\l}$};
\node [poly, fill = bohored] (B2) at (0, -4.5) {$\alpha_{\l}^{(k+1)}$};

\coordinate [] (p1-2) at (0, -2.5);
\coordinate [] (p2-2) at (1.5, -2.5);
\coordinate [] (p3-2) at (1.5, -5.25);

\coordinate [] (f1-2) at (0, -2);
\coordinate [] (f2-2) at (1.65, -2);
\coordinate [] (f3-2) at (1.65, -3.75);
\coordinate [] (f-out) at (1.65, -5.8);

\node [name=res2, draw, circle, minimum size=1pt, inner sep = 0] at(0, -3.75) {$-$};
\node [name=conc2, draw, circle, minimum size=4pt, inner sep = 0, node distance=1cm] at (0, -5.25) {$+$};

\coordinate (out) at (0, -5.8);

\draw[-latex] (p1-2) -- (p2-2) -- (p3-2) -- (conc2);
\draw[-latex, dashed] (f3-1) -- (f2-2) -- (f3-2) -- (res2);
\draw[-latex, dashed] (f3-2) -- (f-out); 

\draw[-latex] (conc) -- (A2);
\draw[-latex] (A2) -- (res2);
\draw[-latex] (res2) -- (B2) node[pos=.35, left] {$\sigma$};
\draw[-latex] (B2) -- (conc2) node[midway, left] {$\sigma$};
\draw[-latex] (conc2) -- (out);

\end{tikzpicture}
        \caption{polynomial}\label{subfig:mgnetp}
    \end{subfigure}
    }
    \hfill 
    \caption{Weight sharing in ResNet and MgNet; (\ref{subfig:resnet}) ResNet-blocks, no weight sharing; (\ref{subfig:mgneti}) MgNet-blocks, shared $A_\l$; (\ref{subfig:mgnet}) MgNet-blocks, shared layers $A_\l$ and $B_\l$ and Poly-MgNet (\cref{subfig:mgnetp}).}\label{fig:resnet_mgnetblocks}
\end{figure}
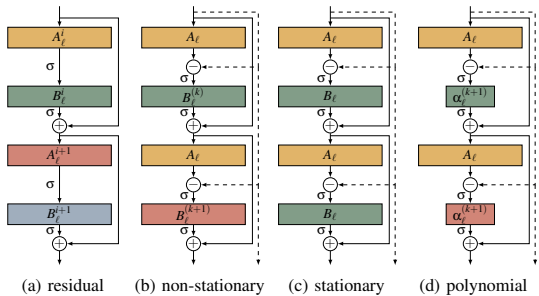

Nevertheless these networks contain $\mathcal{O}(10^7)$ -- $\mathcal{O}(10^{10})$ weights, thus being heavily over parameterized. A reduction of the weight count is clearly desirable, which, however, can result in an undesired bias.
This trade-off is referred to as “bias-complexity trade-off” which constitutes a fundamental problem of machine learning, see e.g.~\cite{shalev-shwartz_understanding_2014}.
In this work, we address this problem from a multigrid (MG) perspective. MG methods are hierarchical solvers for large sparse systems of linear equations that arise from discretizations of partial differential equations~\cite{trottenberg_multigrid_2001}. The main idea of MG consists of two components, namely a local relaxation scheme, which is cheap to apply, but slow to converge as it lacks the possibility to address global features. Thus it is complemented with a coarse grid correction, that exploits a representation of the problem formulation on a coarser scale thus making long range information exchange easier. In classical MG theory this complementarity can be associated with the split of the error into geometrically oscillatory and smooth functions. Where the oscillatory part is quickly dampened by the local relaxation scheme and the smooth part accurately described and dealt with on coarser scales~\cite{trottenberg_multigrid_2001}. The authors of ResNet~\cite{he_deep_2016} already mentioned the inherent similarities between MG and residual layers. This structural connection was further elaborated in~\cite{he_mgnet_2019}, where ResNets, composed of residual layers (representing the smoothers) and pooling operations (representing the coarse grid restrictions), are cast into a full approximation scheme (FAS). The resulting framework, termed MgNet, further exploits the similarity to MG methods, in which the discretized operators stemming from PDEs remain fixed between consecutive coarsenings/poolings. This yields a justification for sharing weight tensors across multiple residual layers, e.g.~\cref{fig:resnet_mgnetblocks}(b) and~\cref{fig:resnet_mgnetblocks}(c), reducing the overall weight count of the network.
A joint perspective of multigrid structures in all dimensions was taken by~\cite{betteray2022}, resulting in an improved weight-accuracy trade-off and thus demonstrates that ResNets are overparameterized.

{In this work, {we exploit} additional structural similarities between MG and residual networks to further reduce the weight count of MgNet.}
{A typical ResNet block is built from two convolutions $A$ and $B$ gated by a residual connection.}
Considering a given system of linear equations $Au=f$, {so-called} \textit{MG smoothing} applies
an operation $B$ to the residual $f-Au$, i.e.,
\begin{equation*}
    u \leftarrow u + B(f-Au),
\end{equation*}
in order to improve $u$, i.e., reduce the residual, iteratively. In a nutshell, this operation has the property of smoothing the error associated to the current approximation $u$. 
In MgNet, $B$ is chosen as a learnable convolution operator, but here we consider a cheaper alternative that reuses the convolution $A$ in its definition. Again motivated by the linear-algebra analogy, we consider polynomials $p(A)$ for $B$ with only a handful of learnable parameters, thus reducing the number of parameters by almost 50\%. A special case and the simplest instance of this idea is the Richardson iteration~\cite{trottenberg_multigrid_2001} where $B=\omega I$, with $I$ the identity matrix and $\omega$ a scalar. Clearly, we do not want to limit ourselves to such trivial polynomials and explore in this work to which extent smoothing iterations using polynomial approximations are suitable within MgNet, e.g.~\cref{fig:resnet_mgnetblocks}(c), to further reduce the number of weights.

We summarize our contribution as follows:
\begin{enumerate}
    \item We introduce a new building block, to further reduce the number of weights in multigrid-inspired CNNs, such as MgNet. These methods are inspired by MG smoothers, namely block smoothers, polynomial smoothers and Richardson smoothers, yielding layer modules of reduced weight counts.
    \item We implement this building block into MgNet and show significant reductions of weight counts while almost maintaining classification accuracy.
    \item Our approach yields further insights into the inherent connection of MG and {ResNets} and demonstrates that MG methodology is useful for the construction of weight-count-efficient CNNs.
\end{enumerate}

The remainder of this article is organized as follows: \Cref{sec:related_works} discusses related works. In \cref{sec:resnet_and_mgnet} we elaborate on the similarities of residual networks and MG, followed by the construction of our layers inspired by MG smoothers. Ultimately numerical results are presented in~\cref{sec:results}.

\section{{Related Works}}\label{sec:related_works}
In this section, we provide an overview of related works, classified into four categories of approaches, sorted from remotely to closely related.

\paragraph{Channel Reduction}
The set of existing methods for the reduction of weight counts and the set of existing methods to reduce the computational complexity of CNNs have a significant intersection.
As one of many possible approaches, a reduction of computational complexity can be achieved by a reduction of the number of channels in convolutional layers. Considering the simplified case where the number of input channels $c$ equals the number of output channels of a given convolutional layer and the filter extent $s$ being equal in both directions, the number of weights of such a layer is given by $s^2 \cdot c^2$. Thus, a reduction of $c$ results in a clear reduction of the number of weights.
In~\cite{molchanov_pruning_2016} it was shown experimentally, that there is redundancy in CNNs, which allows for cutting connections between channels after the CNN has been trained. This process is known as CNN pruning~\cite{NIPS1992_303ed4c6,han_learning_2015,li_pruning_2016}. At the same time pruning and other sparsity enhancing methods~\cite{changpinyo_power_2017,han_dsd_2017} reduce the model complexity in terms of weights. While these approaches first train a CNN to convergence, in~\cite{Gale2019} the CNN is pruned periodically during training. In~\cite{lee2018snip} a saliency criterion to identify structurally important connections is proposed, which allows for pruning before training. We also reduce the weight count before training, however, our approach is based on the inherent similarity of MG and ResNets, utilizing MG methodology to find an explanation for more weight count efficient layer modules.

{A related research area is the field of neural architecture search~\cite{Elsken2019,Cha2022SuperNetIN}, where the goal is to search the space of architectures close to optimal ones w.r.t.\ chosen optimization criteria. If the focus is on computational efficiency, then tuning the channel hyper-parameters is one of many possible approaches~\cite{Gordon2017MorphNetF}, resulting in a reduction of weight count.}
While the aforementioned approaches use optimization procedures to reduce the number of weights, we utilize MG methodology that yields an explanation for the achieved efficiency.

\paragraph{Modified Layers}
Another line of research addresses the specific construction of convolutional layers.
Compared to the aforementioned approaches, the constructions outlined in this section are based on human intuition and conventional methods to improve computational efficiency.
Also in this line of research, there is a close connection between computational efficiency and the reduction of the number of weights. 
One approach to reduce the number of weights is the use of grouped convolutions~\cite{krizhevsky_imagenet_2012,xie_aggregated_2017}, where the channels are grouped into $g$ decoupled subsets, each convoluted with its own set of filters, thus decreasing the weight count to $s^2 \cdot \left(\frac{c}{g}\right)^2$. This decoupled structure impedes the exchange of information across the channels. To overcome this issue, e.g.\ in~\cite{zhang_shufflenet_2018} a combination of groupings and channel shuffling, termed ShuffleNet, has been proposed.

A combination of layers, which also allows for a reduction in floating point operations, are so-called depth-wise separable convolutions, introduced in~\cite{howard_mobilenets_2017,sandler_mobilenetv2_2019,howard_searching_2019} as key feature for the MobileNet architecture: a depth-wise convolution, i.e., grouped convolution with $g=c$ is followed by a point-wise $1 \times 1$ convolution, to create a linear combination of the output of the former. This results in a reduced weight count of $s^2 + 2c$, i.e., each kernel slice of extent $c^2$ is replaced by a rank - one approximation.

\paragraph{Polynomial Neural Networks}
Polynomial neural networks are NNs that produce an output being a polynomial of the input. This approach was {first pursued} via the group method of data handling (GMDH)~\cite{OH2003703}. It determines the structure of a polynomial model {in a combinatorial fashion} and selects the best solution based on an external criterion, e.g.\ least squares. While these methods belong to self-organizing neural networks, another category considers the output of the network as a high-order polynomial of the input~\cite{shin_pisigma_1991,Chrysos_2020_CVPR,Chrysos2022}. Our goal is to utilize polynomials as a building block in CNNs without reframing their purpose to polynomial approximation.
 
\paragraph{Multigrid Inspired Architectures}
In scientific computing, MG methods are algorithms based on hierarchical discretizations, to efficiently solve systems of (non-)linear differential equations (PDEs)~\cite{trottenberg_multigrid_2001,treister_--fly_2011,kahl_adaptive_2018}. Some works utilizes CNNs to solve PDEs~\cite{tomasi2021construction}, e.g.\ by estimating optimal preconditioners~\cite{goetz2018} or prolongation and restriction operators~\cite{katrusa2017}.
Another line of research focuses on architectures based on common computational components as well as similarities of CNNs and MG~\cite{he_deep_2016}.
In~\cite{ke_multigrid_2017} an architecture of pyramid layers of differently scaled convolutional layers, with each pyramid processing coarse and fine grid representations in parallel, is proposed. This MG architecture improves accuracy, while being weight{-}count efficient.

The close similarities between CNNs and MG are further exploited in~\cite{he_mgnet_2019}, where a framework termed MgNet is introduced. It yields a justification for sharing weight tensors within convolutions in subsequent ResNet blocks with the same spatial extent. Utilizing MG in the spatial dimensions, MgNet models have, compared to ResNets with the same number of layers, fewer weights, while maintaining classification accuracy. An alternating stack of MgNet blocks and poolings can be viewed as the left leg of an MG $V$-cycle. Another hierarchical structure, however in the channel dimension, is proposed in~\cite{eliasof_mgic_2020}. Their building block, termed multigrid-in-channels (MGIC), is built on grouped convolutions and coarsening via channel pooling. Utilizing MG in the channel dimensions, this approach improves the scaling of the number of weights with the number of channels from quadratic to linear. A unified MG perspective is taken on both the spatial and channel dimensions in~\cite{betteray2022}. The introduced architecture, called multigrid in all dimensions (MGiaD), improves the trade-off between the number of weights and accuracy via full approximation schemes in the spatial and channel dimensions, including MgNet's weight sharing. Similarly to these approaches, we exploit the inherent similarities between CNNs and MG to further reduce the weight count. Our focus in this work is to cast MG smoothers into CNN layer modules, which has not been studied in related work.

\section{{Residual Networks and Multigrid Methods}}\label{sec:resnet_and_mgnet}
MG consists of two complementary components, the smoother and the coarse grid correction. In a recursive fashion, the coarse grid correction is typically treated by restricting/pooling in spatial dimensions, then smoothing on the next coarser scale and then applying another even coarser coarse grid correction, etc. This structure is already resembled by typical CNNs which alternate between convolutions and poolings.
In MG and in CNNs, the smoother is the main work step with the highest computational effort. A suitable choice of it is vital for the success of an MG method and we will see that these findings are beneficial for CNNs as well.
In this section we introduce {the smoother, the coarsening} and unify the MG and CNN perspective.

\paragraph{Revisiting ResNet and MgNet}\label{subsec:resnet_and_mgnet}
Given a data-feature relation $A(u)=f$, the right-hand-side $f$ represents the data space and $u$ the features. In CNNs, $A$ is learnable and $A(u)=f$ can be optimized~\cite{he_mgnet_2019}. The mappings between data $f \in \R^{m \times n \times c}$ and features $u \in \R^{m \times n \times h}$ are given by
\begin{align}
    A&:  \R^{m \times n \times h} \mapsto  \R^{m \times n \times c}, \;\;\; \text{s.t.\ } A(u) = f \label{eq:A} \\ 
    B&:  \R^{m \times n \times c} \mapsto  \R^{m \times n \times h}, \;\;\; \text{s.t.\ } u \approx B(f), \label{eq:B}
\end{align}
where $m$ and $n$ characterize the spatial resolution dimensions of the input and $c$ and $h$ determine the dimension of the input and output channel respectively, i.e.\ number of input and output channels can differ. $A$ can be considered as a \textit{feature-to-data map}, while $B$ is applied to elements of the data space, is also referred to as \textit{feature extractor}.  In MG the property $u \approx B(f)$ is beneficial for the method's convergence, which will be explained in the following. In the context of CNNs, convolutional mappings usually are combined with non-linear activation functions. For a clear presentation of the similarities between CNNs and MG, 
the non-linearities are omitted for now. Considering a large sparse system of linear equations $A(u)=f$, obtaining the direct solution $u$ is not feasible. Therefore, the true solution is iteratively approximated by $\widetilde{u}$. The resulting error $e = u - \widetilde{u}$ fulfills the residual equation
\begin{equation}\label{eq:residualequation}
    r = f- A (\widetilde{u}) = A(u - \widetilde{u}) = Ae,
\end{equation}
where $r$ is referred to as residual. Now, given an appropriate feature extractor $B$, the approximated error $\widetilde{e} = Br$ can be used to update the approximated solution $\widetilde{u} \leftarrow \widetilde{u} + \widetilde{e}$.
Repeating this scheme with~\cref{eq:residualequation} yields a non-stationary iteration
\begin{equation}\label{eq:iterative_scheme}
   u \leftarrow u + B^{(k)}(f-A(u)) \; \text{ for } k = 1,2, \ldots \
\end{equation}
to solve $A(u)=f$ approximately. Hence, that feature extractors $B^{(k)}$ depend on iteration $k$, yet~\cref{eq:iterative_scheme} can be turned into a stationary scheme with a fixed $B$, i.e.\ a shared weight tensor. On the other hand, interpreting $A$ as a data-feature mapping motivates a fixed $A$, i.e.\ also a shared weight tensor. As examined in~\cite{he_mgnet_2019} the structure of~\cref{eq:iterative_scheme}, with activation functions added, resembles a ResNet-block. Combined with~\cref{eq:residualequation} yields a non-stationary iterative scheme. Explicitly, given the solution $\uk$ of iteration $k$ the residual 
\begin{equation}\label{eq:residualequation_k}
    r^{(k+1)} = f -Au^{(k+1)} = (I - AB^{(k)})r^{(k)}
\end{equation}
is propagated to $k+1$-th iteration and the coefficient $(I-BA)$
equals a ResNet-block $r^{(k+1)} = r^{(k)} + BAr^{(k)}$ (cf. ~\cref{subfig:resnet}). 
The main difference between MgNet and ResNet is sharing one or more weight tensors. Two ResNet-blocks at one resolution level do not share any weight tensors (cf.~\cref{subfig:resnet}). The weight count of such two distinct blocks is given by $4 \cdot (s ^2 \times c \times h)$.
In MgNet, for the non-stationary case, i.e.\ only $A$ is shared, the weight count is reduced to $3 \cdot (s ^2 \times c \times h)$. {For the stationary case, i.e.\ sharing both $A$ and $B$ the weight count for two blocks is only} $2 \cdot (s ^2 \times c \times h)$. The pseudo-inverse of $A$ is generally a dense matrix, which requires the representation as a fully connected weight layer, but this conflicts with the required convolutional structure of the feature extractors $B^{(k)}$ {(respectively $B$)}. Even with an optimal choice of $B^{(k)}$ the convergence of the iteration~\cref{eq:iterative_scheme} is inevitably slow, due to acting locally. However, convolutional operators $B^{(k)}$ are computationally light. A few applications of the iteration have a smoothing effect on features, and the resulting error can be accurately represented at a coarser resolution. 

\paragraph{Resolution Coarsening}\label{subsec:resolution_coarsening}
The restriction of the (residual) data to coarser scales, facilitated by mappings
\begin{equation}
    R_{\l}^{\l+1} : \R^{m_{\l} \times n_{\l} \times c_\l} \mapsto \R^{m_{\l+1} \times n_{\l+1}  \times c_{\l+1}}\ , 
\end{equation}
yields a hierarchy of resolution levels $\l =1, \ldots, L$. On each level $\l$ the smoothing iteration~\eqref{eq:iterative_scheme} can be applied by resolution-wise mappings $A_\l$ and $B_{\l}^{(k)}$, {where on each level $\l$
initially $u_\l^{(0)} = 0$.} {Note that the feature extractors can also be stationary.}
Equivalent to restrictions in MG, in CNNs the resolution dimension is reduced by pooling operations with stride greater than $1$. In CNNs usually the channel dimension is increased in the process. Corresponding to the coarsening leg of a standard MG $V$-cycle~\cite{trottenberg_multigrid_2001}, the combination of $\nu$ smoothing iterations on each resolution level $\l$ and $L-1$ restrictions yields~\cref{alg:coarsening_multigrid}.

\begin{figure}[t]
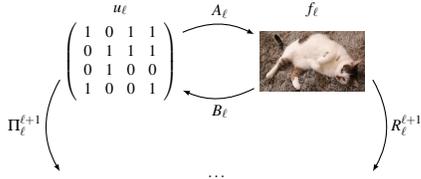

 \centering
  \includestandalone[width=.65\linewidth]{tikz/datafeaturemapping}
 
  \caption{Data-feature relations $A_\l$ and $B_\l$ on resolution level $\l$ followed by transfer to coarser resolution $\l+1$. $A_\l$ applied to the features $u_\l$, calculation of the residual $r_\l = f_\l - A_\l u_\l$, on which the feature extractor $B_\l$ is applied.}
  \label{fig:datafeaturerelations}
\end{figure}

\begin{algorithm}
\DontPrintSemicolon
\SetAlgoLined
\caption{\textbackslash-MgNet($f_\l$)}\label{alg:coarsening_multigrid}
\SetKwInOut{Init}{Initialization}
\Init{$u_{\l} = 0$}\For{$\l = 1, \ldots, L$}{
    \For{$k = 1, \ldots, \nu$}{
        ${u^{(k+1)}_\l = \uk_\l} + B^{(k)}_\l(f_{\l}- A_\l(\uk_{\l}))$\;
        } 
${u^{(0)}_{\l+1}=0}$\;
$f_{\l+1} = R_{\l}^{\l+1}(f_{\l} -A_{\l} {(u^{(\nu)}_{\l}))}$\;
}
\end{algorithm}

\paragraph{Full Approximation Scheme (FAS)}\label{par:fullapproximationscheme}
So far activation functions, potential non-linear poolings and normalization operations, which are characteristic of CNNs, have been disregarded. However, if we now take these into account, we obtain a non-linear overall structure for CNNs. The non-linearity in MG problems yields multiple possible minima, so that the initial solution not only determines the solution, but also has a crucial influence on the convergence rate. Consequently, an initial guess ${u^{(0)}_{\l+1}}$ at the coarser scale determined by the current feature approximation $u_\l$, is likely to dominate ${u^{(0)}_{\l+1}=0}$.
Therefore, the feature approximations of non-linear problems are also projected to the coarser scale by a mapping 
\begin{equation}\label{eq:projection}
     \Pi_{\l}^{\l+1} : \R^{n_{\l} \times m_{\l} \times c_\l} \mapsto \R^{n_{\l+1} \times m_{\l+1} \times c_{\l+1}},
\end{equation} 
to initialize the solution {$u^{(0)}_{\l+1} = \Pi_{\l}^{\l+1}u^{(\nu)}_{\l}$}.
Given this non-trivial initial solution on level $\l+1$, the restricted residual data $f_{\l+1}$ requires an adjustment by $A_{\l+1}(u_{\l +1})$. 

Accordingly, in~\cref{alg:coarsening_multigrid} lines $5$ and $6$ are changed to
\begin{align}
u_{\l+1}^{(0)} &= \Pi_\l^{\l+1}{u^{(\nu)}_\l} \\
f_{\l+1} &= R_\l^{\l+1}(f_{\l} -A_{\l} ({u^{(\nu)}}_{\l})) + A_{\l+1}({u^{(0)}}_{\l +1}).
\end{align}
In the CNN context, the projection operation~\eqref{eq:projection} corresponds to another pooling operation, but it has no exact counterpart in the general ResNet architecture. \Cref{fig:datafeaturerelations} summarizes relevant mappings and schemes the role of $B_\l$ as feature extractor, $A_\l$ as data-feature mapping, followed by restriction and projection, respectively.

\subsection{{Polynomial (smoother) in Residual Blocks}}
Thus far the iterative scheme,~\cref{eq:iterative_scheme}, was either considered to be stationary or non-stationary with matrices $\Bk$ or $B$, respectively. Even though $B$ is a rough and computationally inexpensive approximation for $A^{-1}$, in order to achieve a reduction of the residual we can further reduce the number of learnable weights and increase the interpretability of $B$ by replacing it by a polynomial {(smoother)} $p_d(A) \approx A^{-1}$ of degree $d \in \N_+$, i.e., the residual block then reads
\begin{equation}\label{eq:iteration_with_polynomial}
    u \leftarrow u + p_d(A)(f-A(u)).
\end{equation} 
In here $p_d$ is a polynomial 
\begin{equation}\label{eq:polynom_normalform}
    p_d(A) = \sum_{i=0}^d \alpha_i A^i.
\end{equation} with (learnable) coefficients $\alpha_i$. In the case of polynomial smoothers $(B=\pd(A))$, the non-stationary case is considered, e.g.\ only weight tensor $A$ is shared. Furthermore these models are denoted by MgNet$^{\pd}$. 

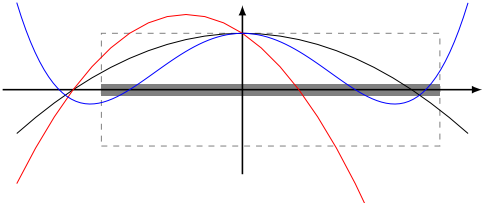
\begin{figure}
\centering
\scalebox{.75}{
 \begin{tikzpicture}
 \begin{scope}
     \clip (-4.25,-2) rectangle (4.25,2);
     \fill[black!50!white] (-2.5,-.1) rectangle (3.5,.1);
     \draw[thin,black!50!white,dashed] (-2.5,-1) rectangle (3.5,1);
     \draw [latex-,thick] (0, 1.5) -- (0, -1.5);   
     \draw [latex-,thick] (4.25,0) -- (-4.25, 0);
     \draw[domain = -4:4, variable = \x]  plot (\x,{(1-(1/3)*\x)*(1+(1/3)*\x)}); 
     \draw[red,domain = -4:4, variable = \x]  plot (\x,{(1-(1)*\x)*(1+(1/3)*\x)});      
     \draw[blue,domain = -4:4, variable = \x,samples=100]  plot (\x,{(1-(1/2)*\x)*(1+(1/3.25)*\x)*(1+(1/2)*\x)*(1-(1/3.25)*\x)});
    \end{scope}
    \end{tikzpicture}}
    \caption{Schematic illustration of polynomials $q_2$, (two blocks), with different choices of roots from an exemplary (real) spectrum.}\label{fig:example_polynomials_real}
\end{figure}

In order to ease the following discussion note that under the assumption that $A$ is diagonalizable, i.e., $A = X\Lambda X^{-1}$ with a matrix $X$ containing the eigenvectors of $A$ as its columns and a diagonal matrix $\Lambda$ of eigenvalues, we find
\begin{equation}
    p_d(A) = \sum_{i=0}^{d}\alpha_i A^{i} = X\left(\sum_{i=0}^{d}\alpha_i \Lambda^{i}\right) X^{-1}.
\end{equation}
Thus the action of the polynomial on a matrix $A$ is determined solely by the evaluation of the scalar polynomial $p_d$ with coefficients $\alpha_i$ on the eigenvalues of $A$. As it alleviates notation we are thus considering the polynomial $p_d$ both as a scalar polynomial and a matrix valued polynomial using the same notation. Moreover, from now on for clarity in notation, {let $\Lambda$ denote the spectrum of $A$.

\paragraph{Residual Blocks as Polynomials}
Let $B = p_d(A)$ be a polynomial feature extractor, than the residual equation~\ref{eq:residualequation} yields 
\begin{equation}\label{eq:residual_factorization}
    r^{(k+1)} = (I- A p_d(A))r^{(k)} = q_{d+1}(A)r^{(0)},
\end{equation}
where the factor $(I-A p_{d}(A))$ itself is a polynomial $ q_{d+1}(A)$ of degree $d+1$. Hence, using~\cref{eq:polynom_normalform} we find 
\begin{equation}\label{eq:qs(A)}
 q_{d+1}(A) = I - Ap_{d}(A) = I - \sum_{i=0}^d \alpha_i A^{i+1}.
\end{equation}
and see that $q_{d+1}$ is normalized with a constant coefficient equal to $1$, i.e., $q_{d+1}(0) = 1$, which implies that residual components belonging to kernel modes of $A$ are unaffected by such a polynomial correction approach\footnote{Clearly such modes are not affected in a general residual block with a (full) parameter matrix $B$ either.}.
Leveraging the normalization of $q_{d+1}$ we obtain its decomposition into a product of linear factors as
\begin{equation}\label{eq:polynomial_q}
   q_{d+1}(A) = \prod_{i=1}^{d+1}(I- \frac{1}{\zeta_i}A) \text{\ with\ } q_{d+1}(\zeta_i) = 0 \text{\ and\ } \zeta_{i} \in \mathbb{C}.
\end{equation}
That is, instead of learning the coefficients $\alpha_{i}$ we can equivalently learn the roots of the polynomial $q_{d+1}$. While it is not immediately clear how $\alpha_i$ need to be chosen to ensure $p_{d}(A) \approx A^{-1}$ to obtain a reduction of the residual, the roots of the polynomial $q_{d+1}$ have an immediate interpretation with respect to the reduction of the residual. Again assuming that $A$ is diagonalizable and taking into account that polynomials tend to be small only close to their roots. That is, writing $r^{(k)} = \sum_{j} \gamma_{j} x_{j}$ as a linear combination of the eigenvectors of $A$ we find
\begin{equation}
    q_{d+1}(A)r^{(k)} = \sum_{j} \prod_{i=1}^{d+1}\left(1-\frac{\lambda_{j}}{\zeta_{i}}\right)\gamma_j x_{j},
\end{equation}
which is small, if the roots $\zeta_{i}$ capture the distribution of the eigenvalues of $A$ correctly. To be more specific it is clear that roots close to the boundaries of 
 the spectrum of A, {$\Lambda$ $\subseteq \mathbb{C}$}, are required in order to prevent catastrophic over correction of the respective eigencomponents of the residual~\cite{saad2013_iterative}(cf.~\cref{fig:example_polynomials_real}). Yet, the aim of minimizing the residual is consistent with keeping the polynomial within the spectrum small. To that end eigenvalues located at the spectrum's boundary are chosen as roots for the polynomial. Obviously $A^T \neq A$ holds, which yields guaranteed complex conjugated pairs of eigenvalues. This can be used to construct a polynomial in linear factor representation, covering the extend of the real axis, e.g.\ \cref{fig:example_polynomials_real}, and quadratic terms, covering the extend of the imaginary area and avoiding complex arithmetic within the CNN at the same time\footnote{E.g.\ nn.ReLU does not support complex values, c.f.\ issues \href{https://github.com/pytorch/pytorch/issues/47052}{\#47052}, \href{https://github.com/pytorch/pytorch/issues/46642}{\#46642}.}.

The residual propagation~\cref{eq:residualequation} for a pair of complex conjugated eigenvalues $z$ and $\bar{z}$ yields
\begin{align}\label{eq:quadratic_residualsol}
       r^{(k+1)}  = (1-\frac{1}{z} A) (1 - \frac{1}{\Bar{z}}A)\rk = \Tilde{q}_{d+1} \rk
\end{align}
where $\Tilde{q}_{d+1}$ itself is a quadratic polynomial of degree $d+1$ with roots $z = a + \mathrm{i}b$ {and its complex conjugated counterpart $\bar{z}$ with $\mathrm{i}$ the imaginary unit.}
Thus, the resulting polynomial is the product of $m$ linear polynomials 
$\hat{q_i}$ and $n$ quadratic polynomials $\Tilde{q_i}$

\begin{equation}\label{eq:polynomial_linear_quadratic}
q_{m +2n}(A) = \prod_{i=1}^m \hat{q}_i (A) \cdot \prod_{j=1}^{n} \Tilde{q}_i (A),
\end{equation}
s.t.\ $m + 2n$.

\begin{figure}
    \centering
    \includegraphics[width=.45\textwidth]{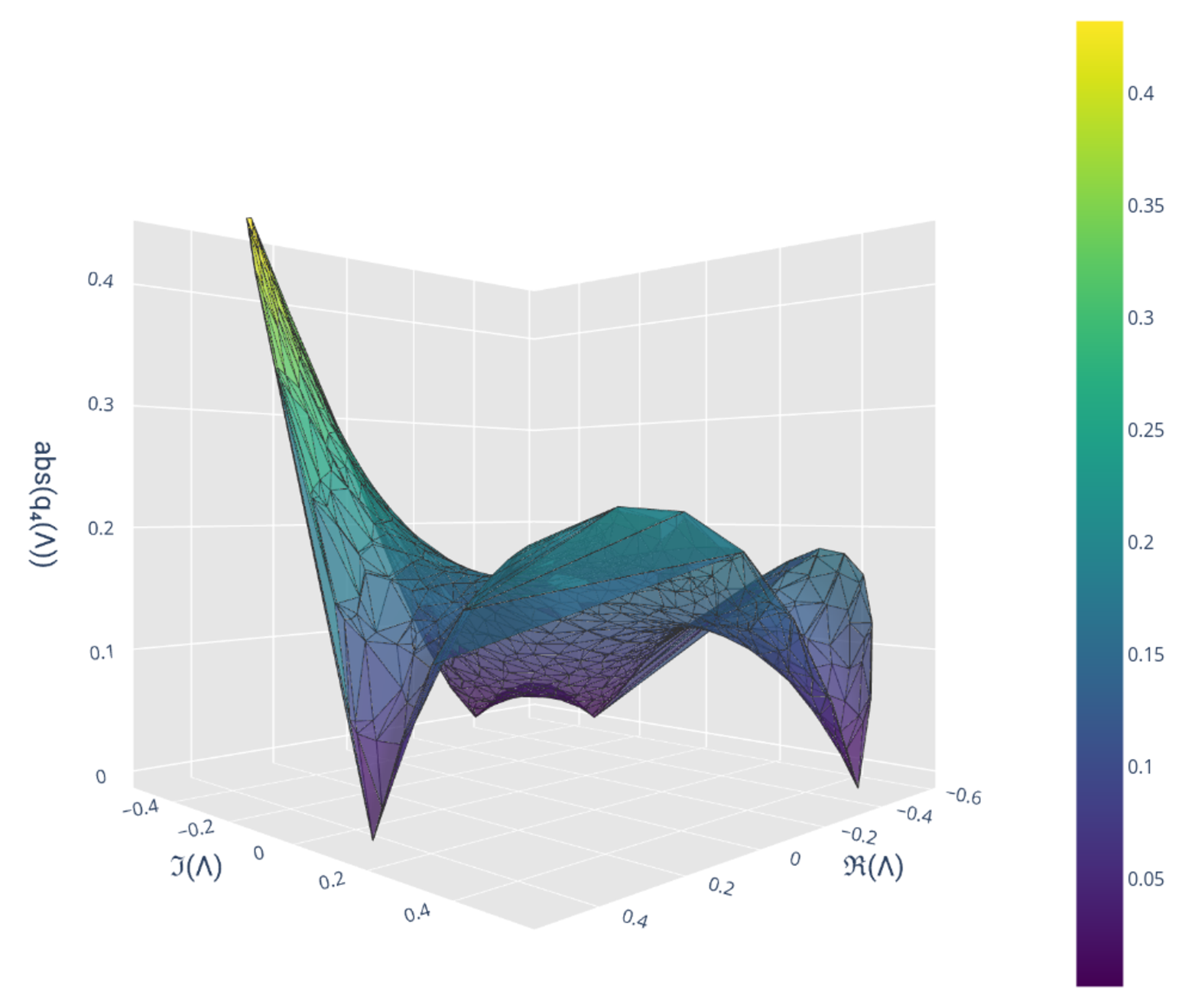}
    \caption{Surface representation of the spectrum $\Lambda$ for the corresponding matrix $A \in \R^{64 \times 64 \times 3 \times 3}$. The $x$-axis represents the real parts, while $y$-axis corresponds to the imaginary part. The $z$-axis illustrates the amplitude of the polynomial function of $\operatorname{abs}(q_4(\Lambda))$, which has roots at the eigenvalues with minimal and maximal real parts, as well as the and complex conjugated pair of eigenvalues with the maximal imaginary part. This visualization allows for an intuitive identification of the spectrum's maximal amplitude.}
    \label{fig:spectrum_evaluated_on_poly}
\end{figure}
The second observation revolves around the fact that the decomposition of $q_{d+1}$ into a product of linear factors can be viewed as a sequence of linear residual blocks, i.e., one residual block with $B = \hat{p}_d(A)$ is equivalent to $d+1$ residual blocks with 
$B^{(k)} = \nicefrac{1}{\zeta_k}$. The quadratic terms also can be viewed as a sequence of pairs of linear residual blocks, i.e.\ two residual blocks with $B=\Hat{q}_{2d}$ are equivalent to $B^{(k)}  = \nicefrac{1}{z}$ and $B^{(k+1)} = \nicefrac{1}{\Bar{z}}$ respectively. The focus of our work is on the polynomial perspective of a residual block, denoted by $q$.
The iteration can be rewritten as
\begin{equation}\label{eq:quadratic_residualequation}
     u^{(k+1)} = \uk + (\frac{2a}{a^2 + b^2} - \frac{1}{a^2 + b^2} )A \rk.
\end{equation}
For the ease of notation from now on the degree of a polynomial $q_d$ is $d$. Consequently the degree of a polynomial $p$ is $d-1$.} Furthermore combining~\cref{eq:iteration_with_polynomial} with~\cref{eq:polynomial_linear_quadratic} results in a polynomial version of MgNet, referred to as Poly-MgNet, which is outlined in~\cref{alg:polyMgNet}.

\begin{algorithm}
\DontPrintSemicolon
\SetAlgoLined

\caption{Poly-\textbackslash-MgNet($f_\l$)}\label{alg:polyMgNet}
\SetKwInOut{Init}{Initialization}
\Init{$u_{1}^{(0)} = 0$}
\For{$\l = 1, \ldots, L$}{
     \For{$k = 1, \ldots, m$}{
     $u^{(k+1)}_\l = \uk_\l + {\frac{1}{\zeta_{\l}^{(k)}}}(f_{\l}- A_\l(\uk_{\l}))$\;
     }
    $u_\l = u^{(m)}$\; 
    \For{{$k = m+1, \ldots, n$}}{
    $z = \zeta_{\l}^{(k)} \in \mathbb{C}$\;
    $a= \Re(\zeta^{(k)}), b = \Im(\zeta^{(k)})$\;
    $\rk = f_\l - A\uk_\l$\;
    $ u^{(k+1)}_\l = \uk_\l + \frac{1}{a^2 + b^2}(2a-A)\rk$\;
    }
$u_{\l+1} = \Pi_\l^{\l+1} u^{(m+n)}_\l$ \;
$f_{\l+1} = R_{\l}^{\l+1}(f_{\l} -A_{\l} (u_{\l})) + A_{\l+1}(u_{\l+1})$\;
}
\end{algorithm}

Due to observations made in our experiments we limit the polynomial with linear factors, i.e.\ $\hat{q_i}$ in~\cref{eq:polynomial_linear_quadratic} to $m=2$. To cover the extent of the spectrum on the real axis we choose the real part ($\Re$) of eigenvalues with smallest and biggest real part for the polynomial roots, i.e.\ ${\zeta^{(1)}} = {\min\Re (\Lambda)}$
and ${\zeta^{(2)}} ={\max \Re(\Lambda)}$ as polynomial coefficients $\alpha^{(k)} = \nicefrac{1}{\zeta^{(k)}}$. Models with polynomial building blocks with a polynomial degree $d=m=2$ are denoted by MgNet$^{q_2}$. For degrees $d \geq 4$ the roots for the quadratic polynomials $\Tilde{q}_i$ are chosen as follows. For a single quadratic term $\Tilde{q}_1$, $z = \operatorname{argmax}(\Im (\Lambda))$ is the eigenvalue with biggest imaginary part ($\Im$), and $\bar{z}$ its complex counterpart. For polynomials of higher degrees we continue to choose roots, that are located at the border of the spectrum to satisfy the requirement of small polynomial values. The roots calculated for a polynomial with degree $d=4$ are deployed in the resulting polynomial $q_4(A)$ followed by its evaluation on $\Lambda$. Consequently, the spectrum is a plane spanned over the roots, e.g.\ shown in~\cref{fig:spectrum_evaluated_on_poly}. The biggest eigenvalue w.r.t.\ the imaginary part, and its complex counterpart are chosen as roots for $q_6(A)$. 
This recursion of spanning the spectrum on the roots to determine new maxima can be repeated as often as required, as summarized in~\cref{alg:roots_higherdegreespolynomials}. The resulting models are denoted by Poly-MgNet$^{q_d}$.

\begin{algorithm}
\DontPrintSemicolon
\SetAlgoLined
\caption{Roots $\zeta^{d}$ for polynomials $d \geq6$}\label{alg:roots_higherdegreespolynomials}
\SetKwInOut{Init}{Initialization}
\Init{$\begin{array}{lcl}\zeta^{(1)} = \max(\Re (\Lambda))\\\zeta^{(2)} =\min(\Re( \Lambda)) \\ \zeta^{(3)} = \operatorname{argmax}(\Im( \Lambda)), & \zeta^{4} = \overline{\zeta^{(3)}}\end{array}$}
\For{$d = 6, 8, \ldots$}{
$q_d = \prod_{k=1}^{d-2}(1-\frac{1}{\zeta^{(k)}}\Lambda)$ \;
$\zeta^{(d-1)} = \argmax ( |q_d (\Lambda)| )$ \;
$\zeta^{(d)} = \overline{\zeta^{(d-1)}}$
}
\end{algorithm}

Another observation is that enhancing the impact of the real valued coefficients by constructing a quadratic version of $\hat{q}$ in~\cref{eq:polynomial_linear_quadratic}, s.t.\
\begin{equation}\label{eq:residual_propagation_A^2}
    r^{(k+1)} = (I - \alpha A^2)\rk = \hat{g}_d(A^2)\rk
\end{equation} has a beneficial impact on accuracy. Omitting iteration indices, a polynomial $\hat{g}(A^2)$ with coefficient $\alpha$ has roots at $\pm \frac{1}{\sqrt{\alpha}}$. Corresponding to the idea to keep the polynomial small within the spectrum we chose the coefficient $\alpha = \frac{1}{\zeta^2}$ with $\zeta_k = \max (|\max (\Re(\Lambda))|, |\min (\Re(\Lambda))| )$ to ensure that the root is at least at the border (or beyond) of the spectrum.
Note that the degree of the overall polynomial $g_d$ is now $d = 2m+2n$. Consequently~\cref{eq:polynomial_linear_quadratic} is rewritten as 
\begin{equation}\label{eq:polynomial_quadratic_quadratic}
g_{2m +2n}(A) = \prod_{i=1}^{m} \hat{g}_i (A^2) \cdot \prod_{j=1}^{n} \Tilde{q}_i (A),
\end{equation}
and models that include the reinforced real part polynomial ${g}$ are denote by Poly-MgNet$^{g_d}$.
The following discussions refer to $\hat{q}_d$, although they can be applied without restriction fo $\hat{g}_d$.
\paragraph{To $\relu$ or not to $\relu$}\label{subsec:reluplacing}
Thus far we ignored the fact, that typically regular ResNet-blocks are build with rectified linear unit ($\relu$) activation functions $\sigma = \max(0, x)$. In ResNet, as depicted in~\cref{fig:resnet_mgnetblocks}, every convolution $A$, $B$ is followed by $\relu$, which also applies for MgNet, independent of shared operations. Furthermore this setting is applied in the special case $B = p_{d-1}(A)$, especially for $d=1$.
The iterative scheme with $\relu$ functions $\sigma$ are given by
\begin{equation}\label{eq:iteration_relu_regular}
    \ukk = \uk + \sigma \, p_{d-1}(A) \, \sigma \,(f- A\uk)
\end{equation}

However, our polynomial view on residual blocks collides with the fact that the $\max$-function typically cannot be a term of a polynomial. Nevertheless $\relu$ has a major influence on a high expressiveness of CNNs which prompts us to review $\relu$ placements in our polynomial residual blocks. To peruse the actual idea of $\relu$-free polynomials a single $\relu$ is applied on every resolution, namely before resolution coarsening level,  
s.t.\ the initial solution is given by
\begin{equation}\label{eq:iteration_relu_outside}
 u^{(0)}_{\l+1}  = \Pi_\l^{\l+1} \sigma\,u^{(\nu)}_\l
\end{equation}
where $\nu = m+n$ blocks denotes a polynomial degree $d=m +2n$. In our experiments we found, that given this setting our models are not able to explain enough non-linearity from the data resulting in a loss in accuracy. To regain expressive capacity of our polynomial models towards data non-linearity, we placed $\relu$ after calculating the residual
\begin{equation}\label{eq:iteration_relu_afterresidual}
    \ukk = \uk + p_{d-1}(A)\,  \sigma \, (f- A\uk),
\end{equation}
which showed to be beneficial to the models performances. Furthermore the combination of~\cref{eq:iteration_relu_outside} and~\cref{eq:iteration_relu_afterresidual} showed to be the most beneficial for the accuracy. {The introduced options are summarized in~\cref{tab:relu_combinations}.}{
\begin{table}[h!]
    \centering
    \scalebox{0.75}{
    \begin{tabular}{c | c c c}
    \hline
        & $ \sigma\, {u^{(\nu)}_\l}$  & $\sigma p_{d-1}(A)$ & $\sigma  r$  \\
         \hline
    \cref{eq:iteration_relu_regular}  &  &  $\times$ &  $\times$  \\
    \cref{eq:iteration_relu_outside} & $\times$  & &  \\
    \cref{eq:iteration_relu_afterresidual}  & & & $\times$ \\  
\cref{eq:iteration_relu_outside} + \cref{eq:iteration_relu_afterresidual} &$\times$ &  &  $\times$\\
    \hline
    \end{tabular}}
    \caption{Overview over possible placements of $\relu$ functions $\sigma$ in a polynomial block $ u \leftarrow u + p_{d-1}(A)r$.}
    \label{tab:relu_combinations}
\end{table}
} Another question of interest is the placement of batch normalization operations ($\bn$) within the polynomials. While their role is not completely clear, we found, that different placements have influence on the classification accuracy of the corresponding model. 
Although $\bn$ is typically applied before $\relu$, it is not required. In the following section, we analyze and discuss various MgNet$^{q_{d}}$ $\bn$ and $\relu$ configurations to identify the most effective setup.

\section{{Experimental Results}}\label{sec:results}
We study the effect of our polynomial building blocks $q_d(A)$ on the accuracy-weight trade-off by evaluating the corresponding models on CIFAR-10.
We report the weight count and mean train and test accuracy with standard deviation (std.) of three runs. Although Poly-MgNet is robust to different coefficient initializations, cf.~\cref{tab_initialization}, we initialize the coefficients of our models based on the spectrum of the convolutions to align with our underlying intuition.
\paragraph{Experimental Setup}
Our models are implemented in Pytorch~\cite{paszke_pytorch_2019}. We train our models in batches of $128$ for $400$ epochs and use an optimizer based on stochastic gradient descent with a momentum of $0.9$ and a weight decay of $10^{-4}$. The initial learning rate is set to $0.05$ which is adapted by a cosine-annealing scheduler~\cite{loshchilov_sgdr_2017}.
\begin{table}[t]
    \centering\scalebox{.75}{
    \begin{tabular}{cc c | c}
     \hline
      Dataset   & polynomial & initialization &  acc ($\pm$ std)   \\
      \hline
      CIFAR-10 & ${q_2}$& $\U(\lmin, \lmax)$ &  $ {94.89} $ ($ 0.16 $) \\  
               & & Xavier  &   $ 94.76 $ ($ 0.16 $)   \\
     & ${g_4}$& $\U(\lmin, \lmax)$  &  ${95.28} $ ($ 0.19 $)   \\
     & & Xavier  &  $ 95.32 $ ($ 0.13 $)  \\
     \hline
    \end{tabular}
    }
    \caption{Influence of coefficient initialization for models Poly-MgNet$^{q_2}$ and Poly-MgNet$^{g_4}$. The coefficients are drawn either uniformly from the spectrum $\Lambda$. s.t.\ $\U(\lmin, \lmax)$ or from $\U(-t, t)$ with $t= \sqrt{ \tfrac{6}{c_{in} + c_{out}}}$. The latter follows the Xavier uniform distribution, which is the standard approach for initializing convolutional weights~\cite{pmlr-v9-glorot10a}.}
    \label{tab:cifar10_init_degrees}\label{tab_initialization}
\end{table}

\paragraph{Weight Count of Residual Blocks}
As the weight tensor $A$ is shared across the polynomial blocks in our approach the only learnable parameter is the single polynomial coefficient 
 $\alpha_\l^{(k)} = \nicefrac{1}{\zeta_\l^{(k)}}$, which is associated to each block. A polynomial of degree $d$ only incurs $d$ parameters and it is thus essentially for free to increase the polynomial degree w.r.t.\ the number of parameters.

More specifically, let $\nu$ be the number of polynomial building blocks on one resolution level $\ell$ and let the size of the weight tensor be given by $s^2 \times c^2$, then the weight count of different types of residual blocks and polynomial setting is given as follows :
\begin{itemize}
    \item ResNet18: $(s ^2 \times c^2) \cdot 2 \cdot \nu$,
    \item MgNet$^{\mathrm{A}}$: $(s ^2 \times c ^2) \cdot (1 + \nu)$,
    \item MgNet$^{\mathrm{AB}}$: $ (s ^2 \times c^2) \cdot 2$, 
    \item MgNet$^{q_d}$: $(s ^2 \times c^2) + \nu$.
\end{itemize}

A regular ResNet18 block is built from $2$ residual blocks on $4$ resolution levels with $[64,128,256,512]$ channels. The overall weight count of ResNet18 is $11.2$ million (M). The corresponding MgNet$\ab$ has the same number of resolution levels, but according to~\cite{he_mgnet_2019}, the number of channels on the fourth layer is reduced to $256$. This setting results in an overall weight count of $2.7$M weights. Poly-MgNet$^{q_d}$ with the same number of resolution levels has a weight count of $1.3$M.
\paragraph{ReLU Placement}

In this section we examine the different options for the placement of activation functions and batch normalizations for polynomials $q_2$, corresponding to~\cref{eq:polynomial_linear_quadratic} and $g_4$, $g_6$ and $g_8$, corresponding to \cref{eq:polynomial_quadratic_quadratic} on CIFAR-10 to determine suitable settings.
\begin{table}[t]
    \centering
    \scalebox{.6}{
    \begin{tabular}{l | r c  r  c r c | c | c}
    \hline
     \multirow{2}{*}{{Model}} & 
       \multirow{2}{*}{\textbf{$\bn$}} & 
       \multirow{2}{*}{$ \sigma\,u^{(\nu)}_\l$} &
       \multirow{2}{*}{$\bn$} &  \multirow{2}{*}{$ \sigma p_{d-1}(A)$}& \multirow{2}{*}{$\bn$} &  \multirow{2}{*}{$\sigma r$} & \multicolumn{2}{c}{accuracy}  \\
    &   &&&&&& test $(\pm \operatorname{std})$ & train \\
         \hline
           & & &  &  &  &  & & \\ [-.9em]  
    MgNet$^{\ab}$ && &   & &  &  &  $95.94$ ($0.27$) & $97.60$  \\
    \hline 
      & & &  &  &  &  & & \\ [-.9em]  
    Poly-MgNet$^{q_2}$    &  $\times$ &  $\times$ & & && & $ 65.27 $ ($ 63.4 $) & $ 64.61 $ \\
    && &  $\times$ & $\times$ & $\times$ & $\times$ & $93.04$  $(0.19)$ & $94.00$ \\
  &  $\times$ & $\times$  &   & &  $\times$  &  $\times$ & $ 94.38 $ ($ 0.35 $)& $ 96.54 $  \\
   &  $\times$ & $\times$  &   & &  $\times$  &   & $ \mathbf{94.89} $ ($ 0.16 $)& $ 97.03 $ \\
    &   & $\times$  & $\times$  & &  $\times$  &   & $ 94.71 $ ($ 0.21 $) &  $ 94.71 $ \\
    \hline 
      & & &  &  &  &  & & \\ [-.9em]  
      Poly-MgNet$^{g_4}$   &  $\times$ &  $\times$ & & & & & $ 94.65 $ ($ 0.33 $)&  $ 96.70 $  \\
      &  &   &  $\times$ & $\times$ &  $\times$  &  $\times$ &  $ \mathbf{95.28} $ ($ 0.19 $) &  $ 97.23 $\\
      &  $\times$ & $\times$  &   & &  $\times$  &  $\times$ &  $ 95.02 $ ($ 0.18 $) & $ 97.13 $\\
&   &  $\times$ & $\times$  & &  $\times$  & $\times$  & $ 95.04 $ ($ 0.09 $)& $ 97.11 $  \\
&   &  $\times$ & $\times$  & &  $\times$  &   & $ 95.00 $ ($ 0.18 $) & $ 97.11 $  \\
    \hline
    \end{tabular}}
    \caption{{Influence of $\relu$ and $\bn$ placement (cf.\cref{tab:relu_combinations}) on the accuracy of Poly-MgNet$^{q_2}$ and Poly-MgNet$^{g_4}$ trained on CIFAR-10. The best accuracy for each model is highlighted in bold. In the blocks corresponding to~\cref{eq:iteration_relu_regular} each $\relu$ is followed by $\bn$ (first row). For Poly-MgNet$^{q_2}$ we consider the naive placement of one $\relu$ before resolution coarsening, i.e.~after the polynomial $q_2$ cf.~\cref{eq:iteration_relu_outside}, we contrast this to an additional $\bn$ after $\relu$ with $\bn$ after the residual $r$ (row 2, 3). For Poly-MgNet$^{g_4}$ we limit the shown results to the one with highest accuracy.  
    }}\label{tab:cifar10_q2_bnrelus}.
\end{table}
In~\cref{tab:relu_combinations} the $\relu$ combinations we determined as the most relevant are summarized. The impact of these combinations on the accuracy of Poly-MgNet$^{q_2}$ are compared in~\cref{tab:cifar10_q2_bnrelus}. Recall that $q_2$ corresponds to $\hat{q}_2$ in \cref{eq:polynomial_linear_quadratic} and
is linear with real-valued roots. We observe that placing $\relu$ and $\bn$ at the same places as in MgNet$\ab$, {i.e.\ both $A$ and $B$ are followed by $\relu$ and $\bn$}, leads to a lower accuracy than MgNet$\ab$. Furthermore the naive approach with a single $\relu$ and no additional $\bn$ (\cref{eq:iteration_relu_outside}) (on one resolution level) is not robust, cf.~\cref{tab:cifar10_q2_bnrelus}. One out of three experimental runs fails to learn, resulting in a low average accuracy with high standard deviation. It is well known that polynomials suffer from instabilities with increasing degree. In our experiments, we observed in general that higher polynomial degrees (greater than eight) often led to instabilities. Polynomials of moderate degree mostly achieved superior accuracy.

In contrast to that, the same $\relu$ placement paired with $\bn$ applied after the calculation of the residual achieves the highest mean accuracy, cf.~\cref{tab:cifar10_q2_bnrelus}. 
Furthermore, we study the $\relu$ placement in Poly-MgNet$^{g_4}$, which is based on polynomials, that are quadratic in the real valued term. Note that $g_4 = \hat{g}_4$ according to~\cref{eq:polynomial_quadratic_quadratic}. 
{This emphasize on the real-valued part}
increases the accuracy by $0.4$ percentage points (pp), cf.~\cref{tab:cifar10_q2_bnrelus}.

{Selected results for the $\relu$ placement in models with polynomials $g_6$ and {$g_8$} are summarized in~\cref{tab:relubnplacing_q4g4}.} 
{Recall, that polynomials with $d>2$, and respectively $d>4$ are composed of a real valued part and an imaginary valued part w.r.t.\ their roots; cf.~\cref{eq:polynomial_linear_quadratic}. Specifically, the composition of $g_6$ is given by $g_6 = \hat{g}_1(A^2) \cdot \hat{g}_2(A^2) \cdot \Tilde{q}_1 (A) $.  For the $\hat{g}$-terms the placement of $\relu$ and bn follows the optimal configuration of $g_4$ determined in~\cref{tab:cifar10_q2_bnrelus}.

We observe that for both Poly-MgNet$^{g_6}$ and Poly-MgNet$^{g_8}$ that the application of $\relu$ and $\bn$ after the residual update are essential for high accuracy. 
}
\begin{table}[t]
    \centering
    \scalebox{.6}{
    \begin{tabular}{c | r c  r  c r c | c | c} 
   \hline
       \multirow{2}{*}{Model} & 
       \multirow{2}{*}{\textbf{$\bn$}} & 
       \multirow{2}{*}{$ \sigma\,u^{(\nu)}_\l$} &
       \multirow{2}{*}{$\bn$} &  \multirow{2}{*}{$ \sigma p_{d-1}(A)$}& \multirow{2}{*}{$\bn$} &  \multirow{2}{*}{$\sigma r$} & \multicolumn{2}{c}{accuracy}  \\
    &   &&&&&& test $(\pm \operatorname{std})$ & train \\
         \hline
 & & &  &  &  &  & & \\ [-.9em]  

  & & &  &  &  &  & & \\ [-.9em]  
  Poly-MgNet$^{g_6}$ &   & &  $\times $ & $\times $   & $\times$  &  $\times$ &   $ 95.25 $ ($ 0.41 $)  &$ 97.29 $ \\
  &     $\times $ & $\times $ & &  & $\times$  &  $\times$  & $ \mathbf{95.55 }$ ($ 0.07 $) & $ 97.34 $ \\
   &   &  $\times $ & $\times $  &  & $\times$  &  $\times$ & $ 95.23 $ ($ 0.13 $) &  $ 97.17 $ \\
  \hline
   & & &  &  &  &  & & \\ [-.9em]  
  Poly-MgNet$^{g_8}$ &  & $\times$ &  $\times $ &    & $\times$  &  $\times$ &   $ 95.33 $ ($ 0.33 $) & $ 97.35 $ \\
   & $\times $ & $\times$ &   &    & $\times$  &  $\times$ &    $ \mathbf{95.60}$ ($ 0.34 $) & $ 97.4 $  \\
\hline
\end{tabular}}\caption{{Influence of $\relu$ and $\bn$ placement in Poly-MgNet$^{g_6}$ and Poly-MgNet$^{g_8}$ 
on the accuracy.}}\label{tab:relubnplacing_q4g4}
\end{table}

\paragraph{Channel Scaling}
\begin{figure}
    \centering
    \resizebox{0.45\textwidth}{!}{\input{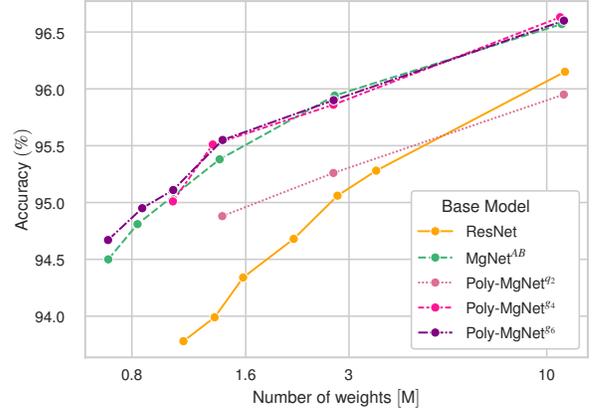}}
    \caption{{Accuracy-weight trade-off of ResNet and MgNet models: influence of overall capacity of residual networks on classification accuracy. The number of channels in the residual blocks of ResNet and MgNet$^{\ab}$ are scaled by {$\nicefrac{1}{\sqrt{2}}$ and $\nicefrac{1}{\sqrt{8}}$}. In contrast, the channels of Poly-MgNet$^{q_d}$ are rescaled by multiplying by $\sqrt{2}$ and $\sqrt{8}$.}}
    \label{fig:scale_resnet_mgnet}
\end{figure}
{After determining feasible settings for the placements of $\relu$ and $\bn$, we study the influence of the capacity in terms of weights on the accuracy of our Poly-MgNet models. {To this end,} we introduce a channel scaling parameter to increase the number of weights in Poly-MgNet according to the weight count of ResNet and MgNet. Respectively, we scale the channels of ResNet and MgNet to meet the capacity of Poly-MgNet. Specifically, the overall initial number of channels\footnote{Initial number of channels in the residual blocks for ResNet18 is $[64, 128, 256, 512]$ and $[64, 128, 256, 256]$ for MgNet} is multiplied by a channel scaling parameter $\lambda < 1$, to reduce the overall weight count to a target limit of $1.3$M weights, which corresponds to the average weight count of polynomial MgNet models.}
{The results are depicted in~\cref{fig:scale_resnet_mgnet}.
A regular ResNet18 has over $11$M weights and achieves an accuracy over $96$\%. Limiting the total weight count to $1. 3$M -- representing a reduction of more than a factor $8$ {--} results in an accuracy drop about $2$ pp. 
In contrast, the standard MgNet$\ab$ with $2.7$M weights already achieves a reduction by a factor of $4$ compared to ResNet18 while maintaining {competitive} accuracy. Despite the lower weight count, MgNet$\ab$ still achieves an accuracy of $96\%$. When the weight count of MgNet is reduced to $1.3$M, we observe a loss in accuracy of $0.6$ pp.
These findings highlight that, compared to ResNet, the weight count can be drastically reduced without significantly affecting the performance in terms of accuracy.}

{Our initial Poly-MgNet utilizes around $1.3$M weights. With this capacity Poly-MgNet$^{q_2}$ achieves an accuracy of $94.89$\%. Compared to ResNet18, which has over $11$M weights, {this} represents a reduction in weight count by a factor of more than $8.5$, {while sacrificing less} than $1.5$ pp of accuracy.
Poly-MgNet$^{g_4}$ and Poly-MgNet$^{g_6}$ achieve accuracies over $95.5$\% with an initial capacity of approximately $1.4$M weights, slightly improving the accuracy of MgNet$\ab$ {by $0.13$ pp while requiring identical capacity}. By increasing the number of channels for both MgNet$\ab$ and Poly-MgNet with building blocks $g_4$ and $g_6$, all models achieve an accuracy of approximately $96.60$\%, thereby drastically improving the accuracy-weight trade-off compared to the standard ResNet18.}

\bibliographystyle{apalike}
{\small
\bibliography{lit, lit2}
}

\end{document}